\documentclass[10pt,twocolumn,letterpaper]{article}

\usepackage{times}
\usepackage{epsfig}
\usepackage{graphicx}
\usepackage{amsmath}
\usepackage{amssymb}
\usepackage{subfig}
\usepackage{placeins}

\usepackage[pagebackref=true,breaklinks=true,letterpaper=true,colorlinks,bookmarks=false]{hyperref}

\begin{document}

\def\dror#1{\textsc{(Dror says: }\marrow\textsf{#1})}
\def\brett#1{\textsc{(Brett says: }\marrow\textsf{#1})}
\def\alex#1{\textsc{(Alex says: }\marrow\textsf{#1})}

\title{Large-Scale 3D Scene Classification With Multi-View Volumetric CNN}

\author{Dror Aiger\\
Google Inc.\\
\\
{\tt\small aigerd@google.com}
\and
Brett Allen\\
Google Inc.\\
\\
{\tt\small brettallen@google.com}
\and
Aleksey Golovinskiy\\
Google Inc.\\
\\
{\tt\small agolovin@google.com}
}

\maketitle

\begin{abstract}
We introduce a method to classify imagery using a convolutional neural network (CNN) on multi-view image projections. The power of our method comes from using projections of multiple images at multiple depth planes near the reconstructed surface. This enables classification of categories whose salient aspect is appearance change under different viewpoints, such as water, trees, and other materials with complex reflection/light response properties. Our method does not require boundary labelling in images and works on pixel-level classification with a small (few pixels) context, which simplifies the creation of a training set. We demonstrate this application on large-scale aerial imagery collections, and extend the per-pixel classification to robustly create a consistent 2D classification which can be used to fill the gaps in non-reconstructible water regions. We also apply our method to classify tree regions. In both cases, the training data can quickly be generated using a small number of manually-created polygons on a map. We show that even with a very simple and standard network our CNN outperforms the state-of-the-art image classification, the Inception-V3 model retrained from a large collection of aerial images.
\end{abstract}


\begin{figure}[hbt]
  \centering
  \subfloat[]{\includegraphics[width=0.21\textwidth]{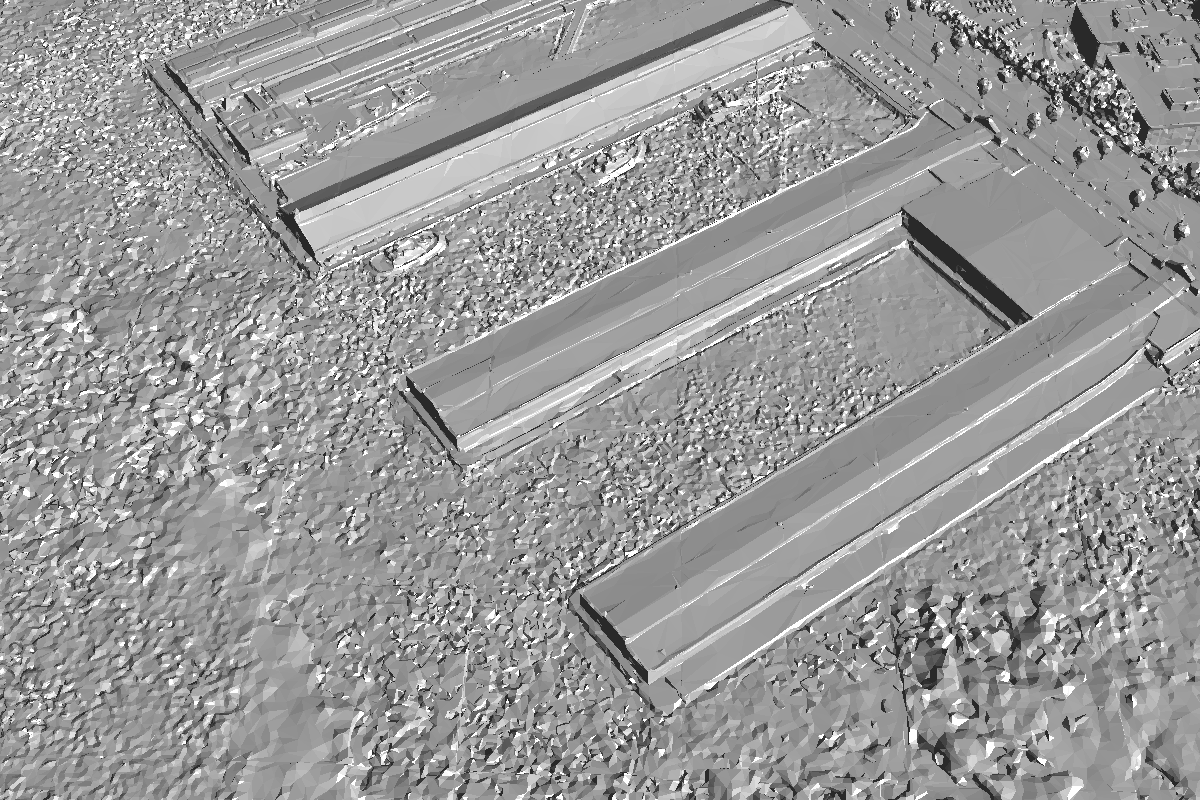}} \quad
  \subfloat[]{\includegraphics[width=0.21\textwidth]{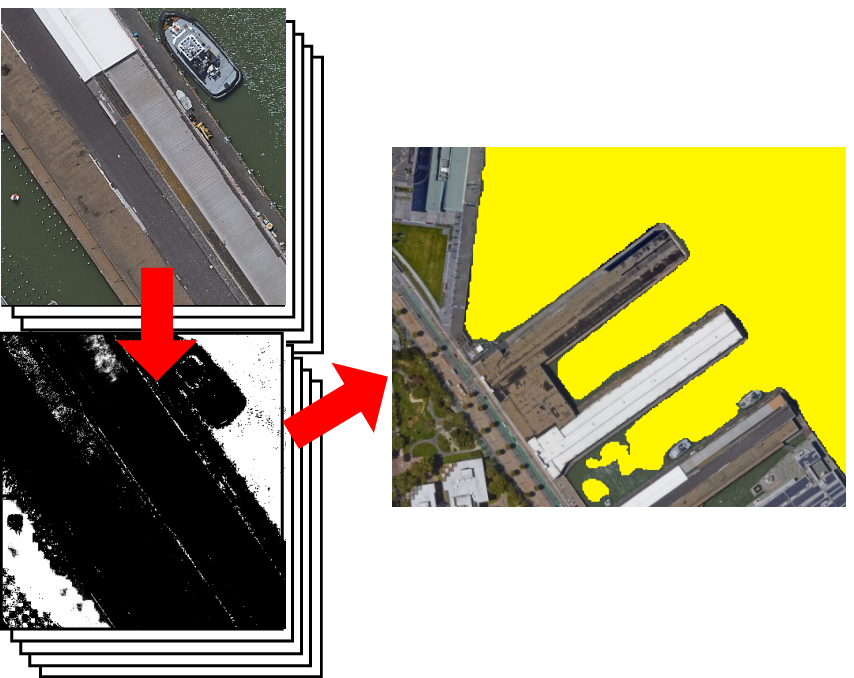}}

  \subfloat[]{\includegraphics[width=0.21\textwidth]{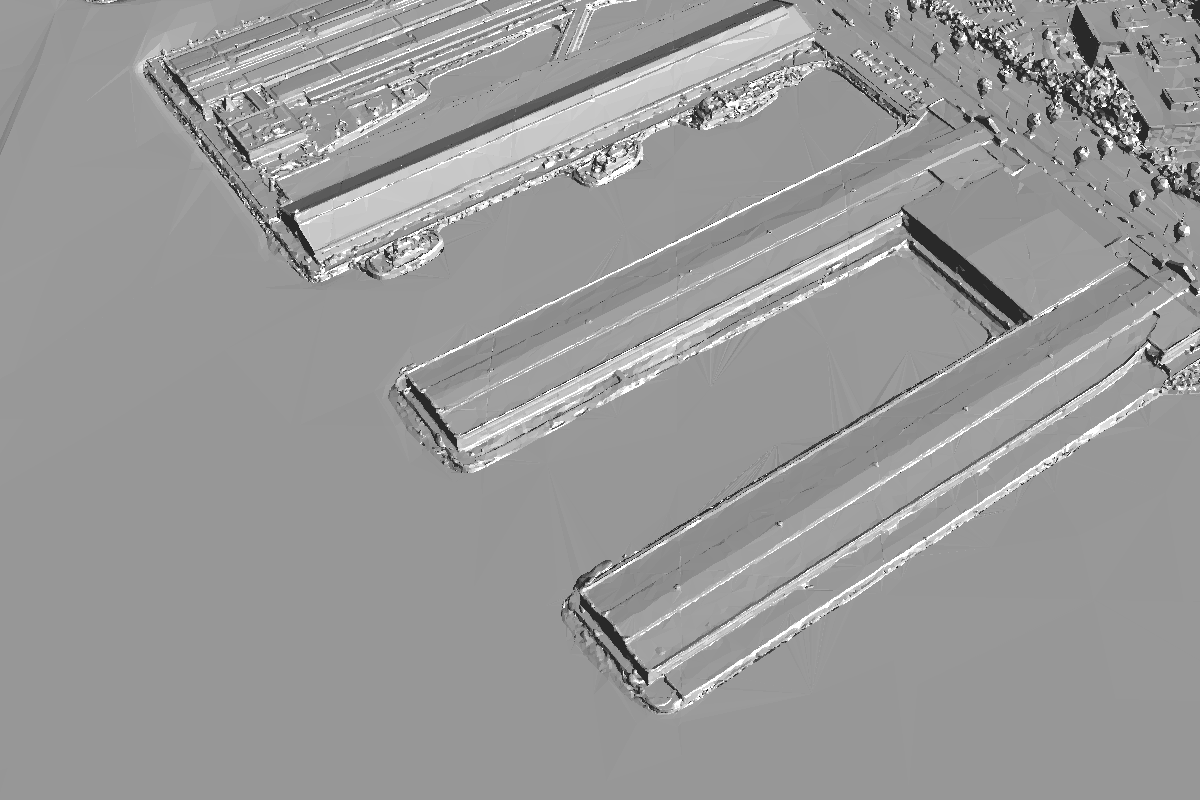}} \quad
  \subfloat[]{\includegraphics[width=0.21\textwidth]{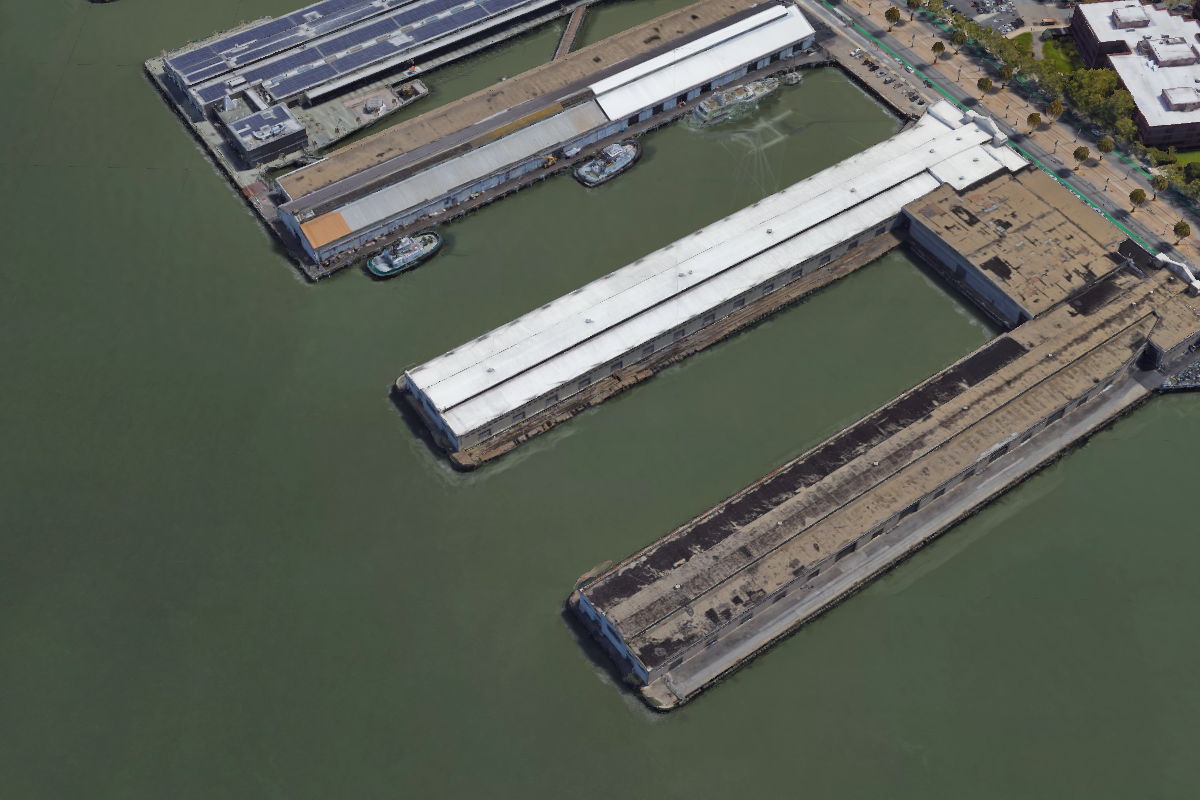}}
  \caption{Aliasing in stereo reconstruction causes water regions to be very noisy (a). We apply our water classification technique to each source image's depth map and its projections, and aggregate the result into a 2D mask (b). Using this mask, we can replace the water with a smooth surface (c) and texture this mesh (d).}
  \label{fig:teaser_result}
\end{figure}

\section{Introduction}

Methods to reconstruct 3D geometry from images have become increasingly effective at creating photorealistic 3D scenes, both in the research community and in a variety of commercial products (an overview of stereo reconstruction methods can be found in \cite{SCDSS06}). Approaches targeting high quality typically use ``dense stereo'' reconstruction. These usually define a ``photoconsistency'' volume as a scalar function on 3D space whose value increases with the similarity of projected images. A high photoconsistency value at a 3D position suggests the existence of a surface. We propose using the projected images that serve as input to such photoconsistency volumes to classify a 3D volume. We further develop a step that merges the classifications of these volumes into a consistent 2D result for the case of aerial imagery.

The following properties are desirable in a practical, large-scale classification system:

\begin{itemize}
\item Scale. It must generalize well, to avoid significantly increasing the size of training set with increased use.
\item Ease of curating / creating training sets. We seek to avoid needing a huge, manually-labelled set of examples as is typical for CNN-based labeling / segmentation methods.
\item Robustness to reconstruction / photoconsistency computation. The system must continue to work as the underlying stereo pipeline changes.
\item Very high accuracy. High numerical classification rates may be misleading, since even a small error rate can lead to a preponderance of visible artifacts.
\end{itemize}

While our approach is general to any volume and set of posed images, we consider it in the context of (a) depthmap creation, in which a source pixel corresponds to an optic ray through the volume, and (b) aerial imagery, where many applications desire as a final result a 2D classified raster. We examine the challenges of such a system and how our approach addresses them in the context of two specific applications. 

First, we target the classification of water. While dense stereo methods can create high-resolution geometry under appropriate circumstances, it is difficult to judge when they fail, and failure leads to egregious errors (see Figure~\ref{fig:teaser_result}(a)). In the case of aerial imagery, the most problematic artifacts in practice come from attempting to reconstruct water, which is tricky for stereo systems because it is often moving and has specular highlights that can create deceptive photoconsistency maxima. Our method can identify and fill such errors, leading to results in Figure~\ref{fig:teaser_result}(c). 

Second, we apply our system to classify trees. Trees are an example of a ubiquitous object useful to classify in a number of applications, ranging from visualization to modeling to GIS.

The problem offers many challenges, which we examine through the lens of the above case studies. One approach might be to explicitly classify source images. Visually, water is deceptively difficult to classify (see Figure~\ref{fig:hard} for a few examples of water vs non-water patches), so a classifier operating explicitly on source images will have a hard time. Our method overcomes this problem by learning from the \textit{interactions} between projected image features rather than just the images. This is similar to how photoconsistency volumes are created, but we avoid having to design the similarity measure and choose parameters such as window sizes.

\begin{figure}[hbt]
  \subfloat[]{\includegraphics[scale=0.3]{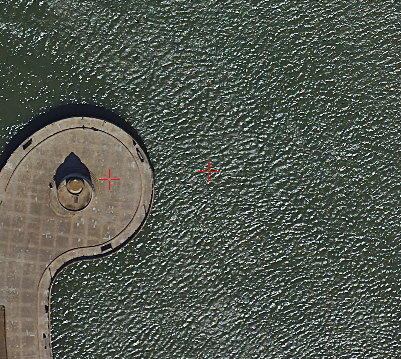}} \hfill
  \subfloat[]{\includegraphics[scale=0.35]{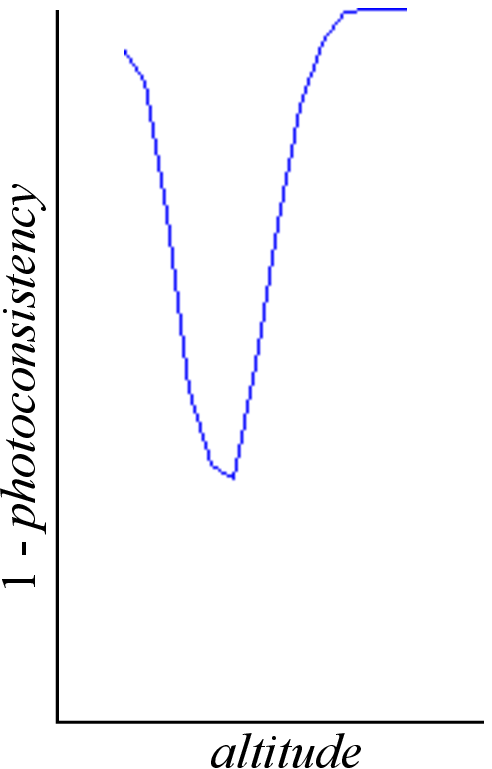}} \hfill
  \subfloat[]{\includegraphics[scale=0.35]{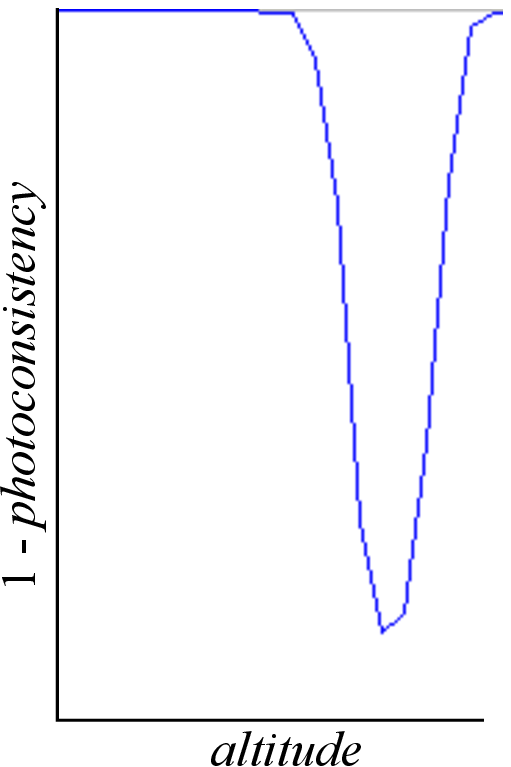}}
  \caption{(a) In wavy water, we can discover image patches with high correlation through pure chance. For example, observe the plane-sweep photoconsistency graphs for the land and water depth slices marked with red crosses are shown in (b) and (c) respectively. The characteristics of these graphs are not sufficient to distinguish water from land.}
  \label{fig:phc_hard}
\end{figure}

An alternative approach to the classification problem would be to create a set of heuristics that operate on the photoconsistency volume. Even ignoring the inherent labor required to make separate heuristics for each classification problem, this is a challenging task for any given particular problem. For example, Figure~\ref{fig:phc_hard} shows graphs of a photoconsistency for a land and a water area, where it happens that the curves are quite similar.

Our method overcomes these challenges. Our contributions include:

\begin{itemize}
\item A CNN-based method to classify the optic rays through a photoconsistency volume. This results in a classified 3D point cloud. Pixel-level correlations among views help classify materials with complex reflection functions, such as water and trees. Using context-based semantic segmentation would require large, fully-labelled image portions in our training set. In contrast, our system is only "weakly" supervised in the sense that we do not mark boundaries between classes in images, needing only sparse pixel labels for its samples.
\item An extension of the above to use results from multiple images to create a consistent 2D mask, aggregating probabilities from all available classified 3D points.
\item For the water classification application: a method to fill the gaps in reconstructed regions that were classified as water including robust estimation of the water level from the boundaries and a filling algorithm.
\end{itemize}

While our system is general, the case study in this paper focuses on water and trees, leaving to future work the exploration of other applications of the multiview pixel-level correlation property. Our main claim is that for many classes of objects, classification based solely on individual images is challenging, especially when practical considerations limit the size of training sets in size and character (we do not assume densely-labelled training sets). We overcome this challenge by using information from multiview correlations, which encode how appearance changes with viewpoints. We also note that for world-wide applications, there are always regions where context cannot help because a pixel does not have any context very far away (e.g. large bodies of water). This means that we must be related on pixel level classification.

\paragraph{Previous work}Our problem is essentially semantic segmentation where we restrict ourselves to using "weakly" supervised learning in the sense that we do not mark boundaries between classes in images. This restriction comes from the desire to perform multi-class segmentation over the entire world; we want to avoid the complex labelling / curation process needed for a training set that accounts for the variations present in the input data. Even for the specific case of water classification, this variation is large and difficult to capture with a small training set. So we prefer to train our algorithm on just a few manually-selected regions around the globe, where each region contains just one class.

Semantic segmentation is commonly handled using a state-of-the-art
convolutional neural network (CNN) framework \cite{LSD15},\cite{DSLD15}. Several papers investigated weakly-supervised learning due to the time consuming requirements (per-pixel labelling) in semantic segmentation \cite{WHC14},\cite{PKD15},\cite{BRFF16}. Semantic segmentation can be achieved by classifying image patches around each input pixel. Convolutional networks are at the core of most state-of-the-art
computer vision solutions for many tasks. Deep convolutional networks are mainstream, yielding very good performance in various benchmarks used at large scale (e.g., ~\cite{SZ14}\cite{SLJSRAEVR15}\cite{KSH12}\cite{ZRS15}). Recently, the Inception-v3 architecture~\cite{SCDSS06} has been shown to achieve very good performance on the ILSVRC 2012 classification challenge with substantial gains over the state of the art. All of these works address image segmentation where the input is just one image. Semantic segmentation needs context, so large parts of images with dense labels are required for training data. Producing such a huge, fully-labelled training set for world-wide classification, with a variety of challenging regions and contexts, is very difficult. Volumetric and multi view CNN classification was previously used in~\cite{RSNDYG16} but the motivation there is object categorization from geometry. We focus on pixel-level classification. As far as we know this is the first attempt to classify image pixels by considering the properties captured by the correspondences among several images. Moreover, the classification in this context leads to a $3D$ point cloud classification. Using pixel-level classification with tiny context (small patches) makes the labelling much easier as we can sample sparse pixels classified as land or water.

In our application we exploit the fact that several images see the same structure and we use plane sweep multi-view stereo and its multi-view structure. We use projections to all available images (or a subset) and let our CNN to learn their mutual correlation. Using the additional information we are able to use a quite standard CNN very much resembling the LeNet CNN~\cite{LBBH98} with weak supervision to obtain the world-wide accurate semantic segmentation that is essential for our $3D$ reconstruction. We are not aware of any prior work that does this but we do our best to compare our work against the state of the art in image-based classification.

A related direction has been investigated by~\cite{hane2013joint}, with the goal of performing joint reconstruction and segmentation. Compared to our approach, their work uses training-derived classification priors: (a) appearance likelihoods derived from individual source images, and (b) geometric priors derived from a 3D dataset. Our volume-based approach expands on (a): we find that classification results are vastly improved by a system that explicitly considers relationships between multiple image projections. We hope to avoid the chicken-and-egg problem of obtaining a 3D training set as required for their system in (b).

\section{The raw volumetric classification}
\subsection{Plane sweep stereo}
\begin{figure}[hbt]
  \centering
\includegraphics[width=0.3\textwidth]{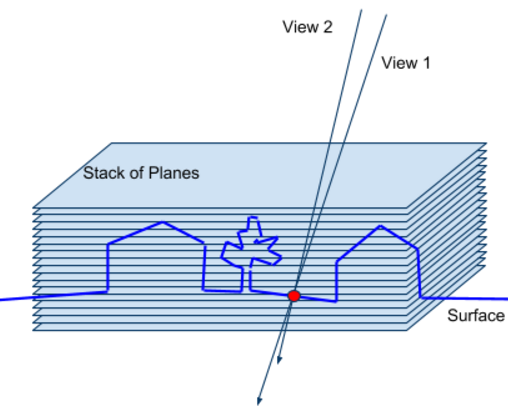}
  \caption{Plane Sweep Stereo}
  \label{fig:pss}
\end{figure}

Our algorithm takes its structure from the $3D$ reconstruction technique referred to as plane sweep stereo. There are two (or more) slightly offset views imaging a scene (Figure~\ref{fig:pss}). Under favorable conditions, a point on the $3D$ surface, when projected to the images, should have the same value. The voxel containing this surface point is said to have high photoconsistency. To help disambiguate among many plausible high-photoconsistency surfaces, a smoothness prior is added. The rough form of the algorithm is then:
\begin{itemize}
\item{Form a voxel grid containing the surface}
\item{Project each voxel to source image and compute a photoconsistency score}
\item{Solve for a depthmap in the voxel grid that has high photoconsistency and high smoothness}
\end{itemize}

General stereo background and alternative methods can be found in~\cite{SCDSS06}. Within their taxonomy, our algorithm can be summarized as:
\begin{itemize}
\item{Scene representation: we use a depth map to represent a scene}
\item{Photo-consistency measure: we use a measure in scene space}
\item{Visibility model: outlier-based. We don’t explicitly model occlusions, and instead assume that lowest $k$ photoconsistency scores might be from occluded views}
\item{Shape prior: In absence of information, we favor horizontal planes (although this can be modified to an extent)}
\item{Reconstruction algorithm: use graph cuts~\cite{BVZ01} to compute optimal depth maps, and merge them in a post-process}
\item{Initialization requirement: we assume that we can initialize a voxel grid that is guaranteed to contain the $3D$ surface}
\end{itemize}

\subsection{Difficulty of image classification for water}

\begin{figure}[hbt]
\captionsetup[subfigure]{labelformat=empty}
  \centering
  \subfloat[]{\includegraphics[width=0.05\textwidth,height=0.05\textheight]{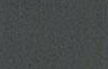}} \thinspace
  \subfloat[]{\includegraphics[width=0.05\textwidth,,height=0.05\textheight]{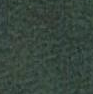}} \thinspace
  \subfloat[]{\includegraphics[width=0.05\textwidth,,height=0.05\textheight]{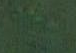}} \thinspace
  \subfloat[]{\includegraphics[width=0.05\textwidth,,height=0.05\textheight]{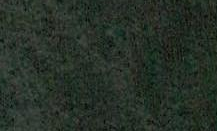}} \thinspace
  \subfloat[]{\includegraphics[width=0.05\textwidth,,height=0.05\textheight]{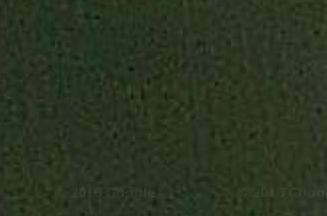}} \thinspace
  \subfloat[]{\includegraphics[width=0.05\textwidth,,height=0.05\textheight]{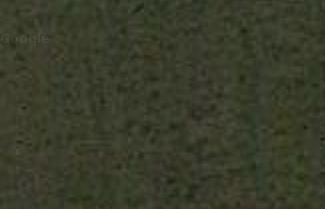}} \thinspace
  \subfloat[]{\includegraphics[width=0.05\textwidth,,height=0.05\textheight]{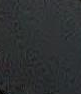}}

  \subfloat[]{\includegraphics[width=0.05\textwidth,height=0.05\textheight]{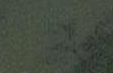}} \thinspace
  \subfloat[]{\includegraphics[width=0.05\textwidth,,height=0.05\textheight]{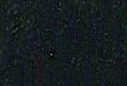}} \thinspace
  \subfloat[]{\includegraphics[width=0.05\textwidth,,height=0.05\textheight]{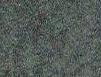}} \thinspace
  \subfloat[]{\includegraphics[width=0.05\textwidth,,height=0.05\textheight]{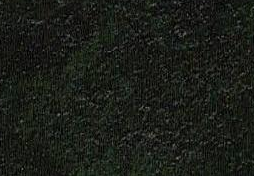}} \thinspace
  \subfloat[]{\includegraphics[width=0.05\textwidth,,height=0.05\textheight]{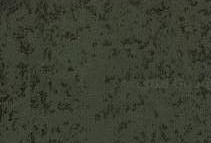}} \thinspace
  \subfloat[]{\includegraphics[width=0.05\textwidth,,height=0.05\textheight]{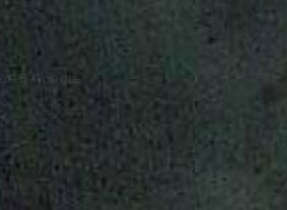}} \thinspace
  \subfloat[]{\includegraphics[width=0.05\textwidth,,height=0.05\textheight]{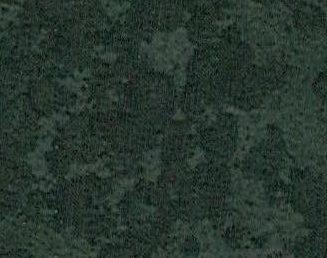}}

  \caption{It is hard to distinguish water and land just from image patches. Can we tell which is water, which is land (top is land, bottom is water)?}
  \label{fig:hard}
\end{figure}

To motivate our approach, in Figure \ref{fig:hard} we show 14 image patches; 7 are water and 7 are land. It is clear that distinguishing based on these patches alone is quite hard. Our $3D$ classification makes use of multi-view image projections and analyzes not only source image patches, but the \emph{interactions} between projections of multiple images. This gives our method more power than image classification alone.

\subsection{Multi-view volumetric data representation}
Our classification is done in $3D$. Using the plane sweep algorithm we obtain for each view (per each input image) a surface which is the computed depthmap corresponding to this view. Our goal is to classify each pixel in this depthmap (a point in $3$-space). In a subsequent section, we will use all depthmaps together to compute a robust $2D$ mask. To take into account information from all views and to account for spatial arrangement in the $3D$ space we represent the data as a sub-volume centred at each $3D$ location on the computed depthmap. This enables us to adapt state-of-the-art CNN-based classifiers designed for image classification, extending them to $3D$ convolution in a straightforward way. 

\begin{figure}[hbt]
\centering
\includegraphics[scale=0.3]{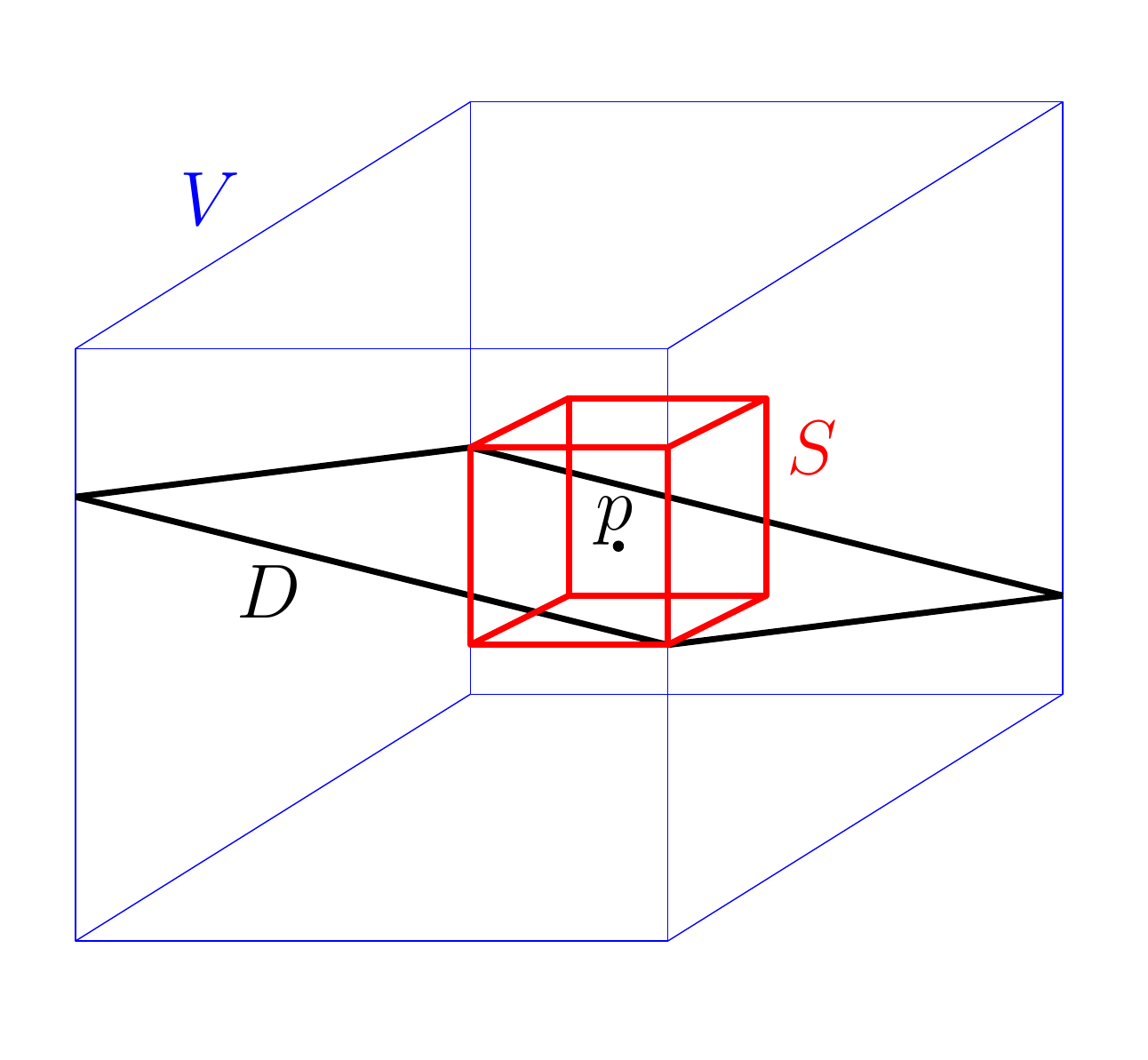}
\caption{A sub-volume extracted around a point $p$ in $3$-space. $V$ is the multi-view volume with projected colors per voxel. $D$ is a depthmap computed by the plane sweep and $S$ is the created sub-volume.}
\label{fig:volume}
\end{figure}

\paragraph{Extracting a set of colors for every cell in a volume grid} We create a $3$-dimensional grid volume $V$ similar to the one used by the plane sweep. Figure \ref{fig:volume} shows the arrangement around a point $p$ in the volume $V$. For each cell $p$ in this volume, we use the projection matrices corresponding to all input images seeing this region and we store a set of colors (intensity) from each projected pixels. Our volume therefore contains $k$ colors (one for each image) for every $3D$ grid cell. For time performance and efficiency our plane-sweep algorithm works in a coarse-to-fine manner, so we actually compute the projections only for points in the sub-volume around each depthmap pixel. For faster runtime, we use grayscale intensities rather than full RGB color.

\paragraph{Creating the sub-volumes for classification} Let $D$ be a depthmap corresponding to one of the views. $D$ is embedded in $3$ space so we can take the set of projected colors from $V$ for every point around any depthmap pixel in $D$. For classification and training we extract a sub-volume around every pixel in $D$ containing all grid volume cells from $V$ around the surface point. Having $k$ images per volume grid cell, we arrange our data in a $3D$ array with one color per entry, containing the $k$ samples in the $z$ axis of the volume. As commonly done for images, we normalize the colors in the sub-volume to compensate for lighting variations regularly observed among all input images.

\begin{figure}[hbt]
\captionsetup[subfigure]{labelformat=empty}
  \centering
  \subfloat[]{\includegraphics[width=0.05\textwidth,height=0.05\textheight]{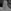}} \thinspace
  \subfloat[]{\includegraphics[width=0.05\textwidth,height=0.05\textheight]{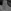}} \thinspace
  \subfloat[]{\includegraphics[width=0.05\textwidth,height=0.05\textheight]{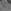}} \thinspace
  \subfloat[]{\includegraphics[width=0.05\textwidth,height=0.05\textheight]{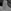}} \thinspace
  \subfloat[]{\includegraphics[width=0.05\textwidth,height=0.05\textheight]{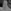}} \thinspace
  \subfloat[]{\includegraphics[width=0.05\textwidth,height=0.05\textheight]{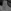}} \thinspace
  \subfloat[]{\includegraphics[width=0.05\textwidth,height=0.05\textheight]{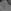}} \thinspace
  \subfloat[]{\includegraphics[width=0.05\textwidth,height=0.05\textheight]{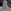}}

  \subfloat[]{\includegraphics[width=0.05\textwidth,height=0.05\textheight]{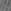}} \thinspace
  \subfloat[]{\includegraphics[width=0.05\textwidth,height=0.05\textheight]{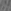}} \thinspace
  \subfloat[]{\includegraphics[width=0.05\textwidth,height=0.05\textheight]{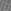}} \thinspace
  \subfloat[]{\includegraphics[width=0.05\textwidth,height=0.05\textheight]{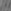}} \thinspace
  \subfloat[]{\includegraphics[width=0.05\textwidth,height=0.05\textheight]{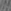}} \thinspace
  \subfloat[]{\includegraphics[width=0.05\textwidth,height=0.05\textheight]{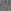}} \thinspace
  \subfloat[]{\includegraphics[width=0.05\textwidth,height=0.05\textheight]{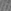}} \thinspace
  \subfloat[]{\includegraphics[width=0.05\textwidth,height=0.05\textheight]{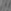}}

  \caption{Data representation. 8 slices in the input vector for water (bottom) and land (top).}
  \label{fig:data_example}
\end{figure}

In our work we use $8$ image projections per grid cell and the sub-volume size we extract around any cell has size $2\times13\times13$. This means that the dimensionality of the data is $2\times8\times13\times13=2704$. As we show in our results, the overall numerical classification over the evaluation data as well as the output $2D$ masks are very good. 

\subsection{The network}
\begin{figure}[hbt]
\centering
\includegraphics[scale=0.4]{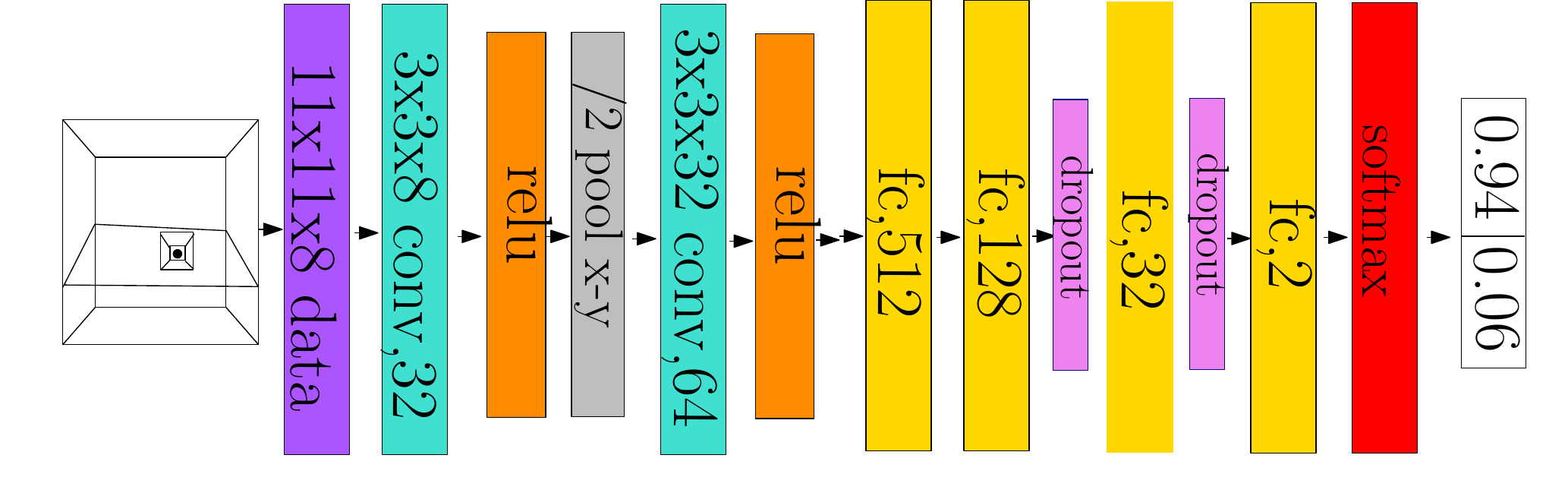}
\caption{The network}
\label{fig:network}
\end{figure}

The input to our network is structured as an $13\times13\times16$ voxel volume while the feature representation is a $512$ dimensional vector. Following several similar ConvNet architecture for classifying $2D$ images and local image patches \cite{LBBH98},\cite{BL14}, the network consists of two convolutional layers with ReLU non-linearity and pooling layers. The volume size is small so we cannot use many pooling layers. Our convolution uses $3D$ kernels. The detailed kernel sizes and number of filters are shown in Figure \ref{fig:network}. We then have $4$ fully connected layers with dropout that prevents the network from over-fitting~\cite{SHK14}. Our loss is the sparse softmax cross entropy between logits and labels and we apply exponential decay to the learning rate. For optimization we use the momentum algorithm \cite{N99}. The implementation of the CNN is TensorFlow~\cite{GOOGLE16}.

\subsection{Training data and numerical results}
\begin{figure}[hbt]
\centering
\includegraphics[scale=0.25]{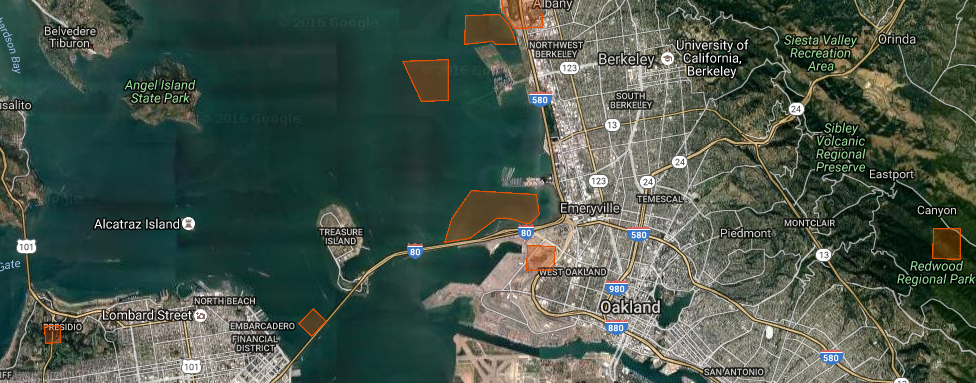}
\caption{Example of training data sampled from San Francisco area}
\label{fig:sf_polygons}
\end{figure}

The training data is considered to be only "weakly" supervised in the sense that we don't mark boundaries in images. We have a full labelling of the training data but the labelling is sparse (per sampled pixel) and the fraction of the training data is very small relative to the size of the data we process (the entire world) and the variations in the input. Our training data consists of small regions defined by ~$100$ polygons of area approximately 1~km$^2$ each, balanced between the classes. The number of regions around the globe we take is fairly small, $30$ different regions overall. The generalization of our classifier must be powerful so these regions would be sufficient to classify all unseen regions. We show in our results that we obtain very good classification in unseen regions with large variations in the classes. An example of some polygons in San Francisco is shown in Figure \ref{fig:sf_polygons}. We sample uniformly the data in the polygons which makes our entire data set consisting of approximately $1M$ vectors for each class.

Note that while the training data is fully labelled, we're able to create the training set in practice by drawing polygons that completely contain a target class. Each such polygon labels all the depthmaps that are located inside it, avoiding the need to perform manual segmentation. One of our pleasant surprises was that including regions containing class boundaries (which would require extensive manual work) is not required to achieve good results. 

We train the system using the described network, randomly dividing the training set to $9/10$ for training and $1/10$ for validation (unseen). 

\section{Generating the $2D$ mask}
\label{sec:mask}
The volumetric classification labels each $3D$ depthmap point with a probability corresponding to each class. One option to use this data is to create a fully-classified $3D$ surface. A simpler option and the one we take here (we left the $3D$ classification for future work) is to create a $2D$ mask (which can be considered as the projection of our points on earth) containing the labels of the class with the largest probability. In the example of water, this yields a boolean mask telling us for each $2D$ point whether it is ``water'' or ``land.'' To do this we have to consider the projection of the points along with their class confidence. We use a method to combine individual probabilities derived directly from Bayes rule~\cite{RN10}. For ease of notation, we consider two classes below in the context of water vs land classification, but this generalizes easily into multiple classes.

The formula is valid only if the probabilities are independent, which is not the case here here, but it is a useful idealization since the statistical correlations between individual probabilities is not known. A similar approach was taken in filtering spam messages~\cite{MAP06}:

$$p=\frac{p_1p_2...p_n}{p_1p_2...p_n+(1-p_1)(1-p_2)...(1-p_n)}$$

Where $p$ is the probability that the projected point is ``water'' and $p_i$ are all individual probabilities of being ``water'' from all depthmaps. This enables us to create a discrete $2D$ mask (a boolean image) containing the label derived by thresholding the projected confidence at $0.5$. To do this, we discretize the $3$-space, insert all points to their corresponding $3D$ cells and then consider a vertical column in this volume as the projection to earth. We then accumulate all probabilities in the columns cells using the formula above. The $2D$ boolean mask is then used to apply further processing steps. For example, for water classification we have to determine the level of the water according to the ($3D$) boundaries of the mask and then fill an artificial surface representing the water to allow a full $3D$ representation of the entire scene containing water and land.

\section{Robustly estimating the water level}

Having obtained a $2D$ mask of regions containing water, we can now remove the noisy reconstructed surface from the water region. Once we remove the noisy surface, we still need to fill in this region with an estimated water surface. Since the altitude of the water itself cannot be obtained from the stereo reconstruction, we instead estimate the altitude of the shoreline, and assume that a local body of water is planar.

\begin{figure}[hbt]
\centering
\includegraphics[scale=0.25]{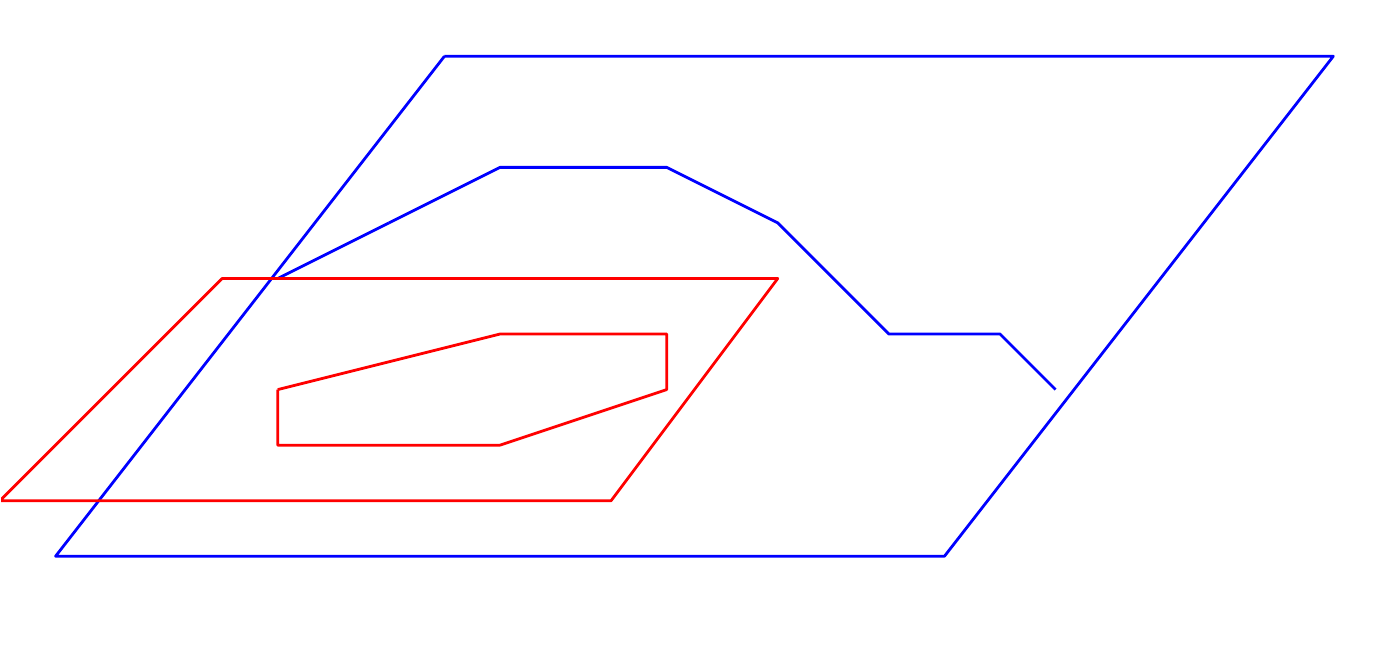}
\caption{Water level estimation using plane fitting on two nearby different connected components.}
\label{fig:fitting}
\end{figure}

Planarity assumption for a local body of water is reasonable -- in fact the water should be horizontal for region of no flow. To determine the shoreline altitude, we use the $2D$ mask obtained by Section~\ref{sec:mask} and we collect samples from all depth map points whose horizontal position is within $40$~m of the shoreline. We group the samples into connected components of shoreline pieces. For each component we fit a plane using RANSAC that robustly represents the most probable water level on the bounds of the water region. Since our system works on a huge scale, we do this step locally in each tile separately. In our fitting process we use a (tiny) prior that prefers levels close to zero because most of the water sources we consider are indeed horizontal (e.g., seas, oceans). Figure~\ref{fig:fitting} shows plane fitting applied to different nearby shorelines, each is a separate connected component that has different water level.

The final water levels are set by smoothly filling in from the shoreline using Laplace's equation. Note that while water is generally flat, in some cases such as rivers it may change altitude over large distances. We handle this situation by running our fitting operation on large ($1.2$~km) tiles that overlap and feathering the margins.

\section{Results}
We present three sets of results. First, we compare our approach with a traditional, image-based CNN, in the context of water classification. Second, we present the end-to-end results of using our system to perform volumetric classification of water, to form the 2D mask, and to fill missing values. Finally, we show the generality of our system by performing tree classification.

\subsection{Comparison with image based Inception-V3 and the power of multiview correspondence}

A natural question about our system is: how much does the volumetric and multi-view nature of the setup really help compared to standard image classification? The patches shown in Figure~\ref{fig:hard} suggest that distinguishing between land and water patches solely by using image data is difficult if not impossible. Using larger context, labelling images with the boundaries as in several semantic segmentation works can definitely help but it is not practical in huge training sets that must be labelled. As mentioned above, context is not always achievable in world-wide applications. It is intuitive then, that the coherency and relationship between corresponding views is valuable. But we wanted to test this intuition.

We applied direct comparisons with the state of the art image classification network -- the Inception-v3 model~\cite{SVISW15}. Of course, one can claim that we do have the poses of the images so in principle, pixel-level correspondence is possible -- indeed, this is part of our contribution. Moreover, along with the pixel correspondences we also apply 3D convolution to exploit these correspondences better. A fair comparison would then be to use exactly the same training data with the same amount of information but with no correspondence, replacing our volume with a set of images (as projected to all images) given to the Inception-V3. This verifies how much information is captured in the correspondence and correlation between the projections. We use the greyscale values of the source image at the depthmap points. Colors could be used as well with increasing time. 

We conducted three experiments to analyse the effect of multiview correspondence which is the core of our method. In all experiments we have used the same amount of training data but with different degrees of correspondence. As mentioned, the core data for our 3D CNN is a volume in space where we project every point to all available images getting grayscale for each such projection. The volume has space size 13x13x2 (for a 25cm grid) and it is centred at each depth point on the surface computed by our 3D plane-sweep reconstruction. Each cell is projected to 8 available cameras so the overall size of the volume is 13x13x16. The three experiments are as follows:

\begin{itemize}
\item Unrelated patches with Inception-v3: We take every z-slize in the volume and create a 13x13 patch out of it. The training set contains all of these patches. The term unrelated means that they are not connected to each other in anyway and each one of them is an independent patch. We indeed expect this to be very hard with no context and no multivire correspondence.
\item Patch correspondence: As above but here we create a new image built from 4x4 such small patches. The training data thus contains images of size 52x52, each of which is a concatenation of 4x4 small 13x13 patches.
\item Full pixel correspondence: This is our 3D volumetric classification having the original volume as our training vectors. The volume creates pixel correspondence as corresponding pixels lie in the same volume column and we apply 3D convolution.
\end{itemize} 

\begin{figure}
  \centering
  \subfloat[]{\includegraphics[width=0.3\textwidth]{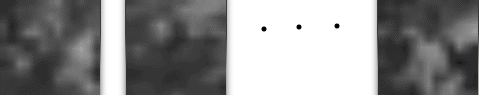}}\hfill
  \subfloat[]{\includegraphics[width=0.1\textwidth]{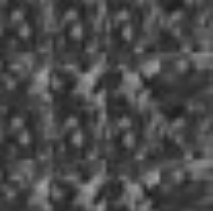}}
  \subfloat[]{\includegraphics[width=0.05\textwidth]{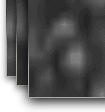}}
  \caption{The structure of the training data in all 3 experiments: (a) No correspondence - Unrelated 16 images per each volume. (b) Patch correspondence - One image containing 16 13x13 patches. (c) Pixel correspondence - 3D volume}
  \label{fig:comp_data}
\end{figure}

Figure~\ref{fig:comp_data} shows the three options of creating training data. The data set for this comparison is composed of 20,000 samples from our entire set. We also take another $1/10$ for test which is used only at the end of the training. Note that here we have two classes and the error weight for each is the same so accuracy is sufficient to evaluate the performance and no precision/recall evaluation is needed. Our supplemental material shows a complete large scale reconstruction of land and water with difficult appearances. The ability to create smooth water regions in the final textured model comes from the large-scale error free water classification. Based on our classification we replace the noisy water with a smooth surface.

\begin{figure}
  \centering
  \subfloat[Train accuracy]{\includegraphics[width=0.13\textwidth]{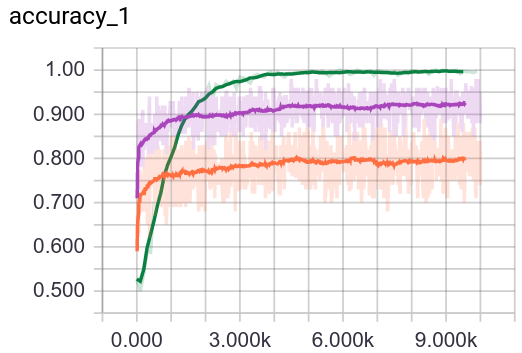}} 
  \subfloat{\includegraphics[width=0.12\textwidth]{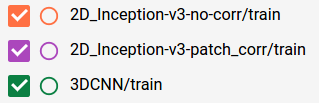}} 
  \subfloat[Validation accuracy]{\includegraphics[width=0.13\textwidth]{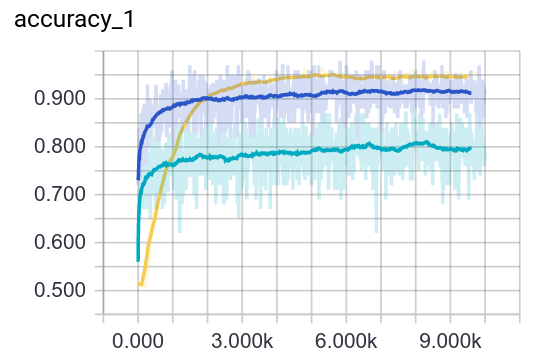}} 
  \subfloat{\includegraphics[width=0.12\textwidth]{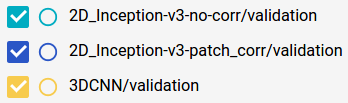}} 
  \caption{Train and Validation accuracy of the 3 experiments. The test accuracy is 79.2 for no correspondence, 90.2 with patch correspondence and 96.3 with 3D CNN.}
  \label{fig:comp}
\end{figure}

A comparison of the 3 experiments can be viewed in Figure~\ref{fig:comp}. We show training accuracy and validation accuracy.

\subsection{Large-scale water classification}

\begin{figure*}
  \centering
  \subfloat[]{\includegraphics[height=1.3in]{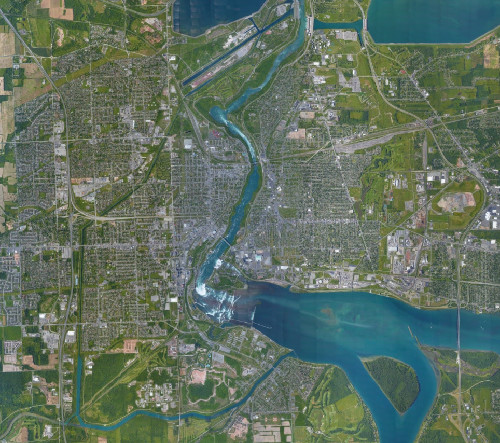}} \hfill
  \subfloat[]{\includegraphics[height=1.3in]{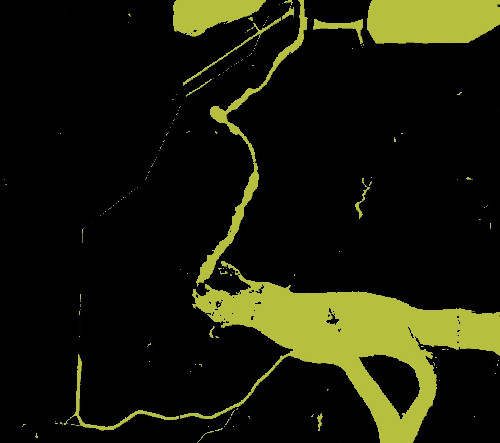}} \hfill
  \subfloat[]{\includegraphics[height=1.3in]{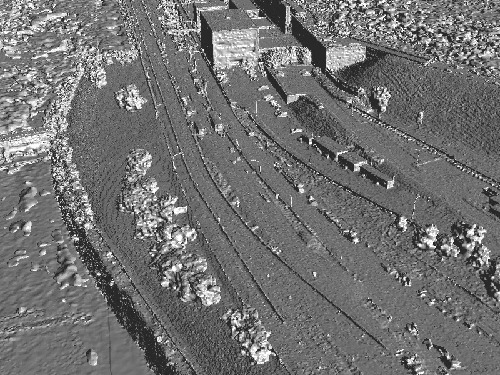}} \hfill
  \subfloat[]{\includegraphics[height=1.3in]{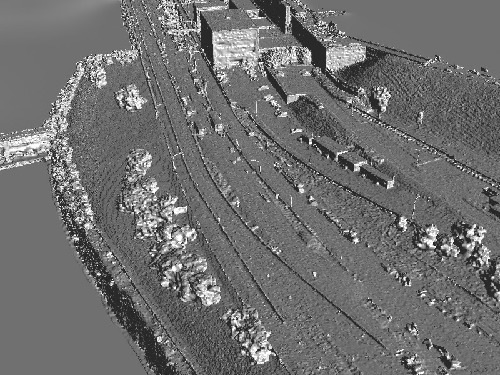}}  
  \caption{(a) A set of imagery covering a 14~km wide region near Niagara Falls, with a large variety of different water bodies. (b) 2D water classification mask for this area aggregated from individual classifications of thousands of images (yellow = water, black = land). (c) Raw 3D reconstruction of a portion of this area. (d) 3D reconstruction after applying our water mask and water level estimation.}
  \label{fig:results_niagara}
\end{figure*}

\begin{figure*}
  \centering
  \subfloat[]{\includegraphics[height=1.1in]{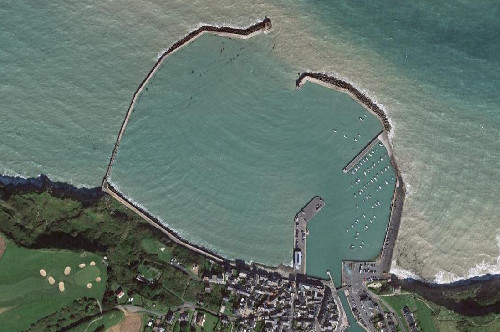}} \hfill
  \subfloat[]{\includegraphics[height=1.1in]{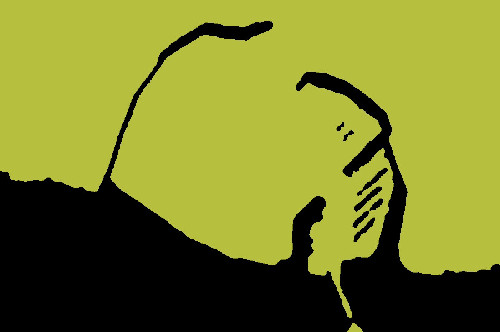}} \hfill
  \subfloat[]{\includegraphics[height=1.1in]{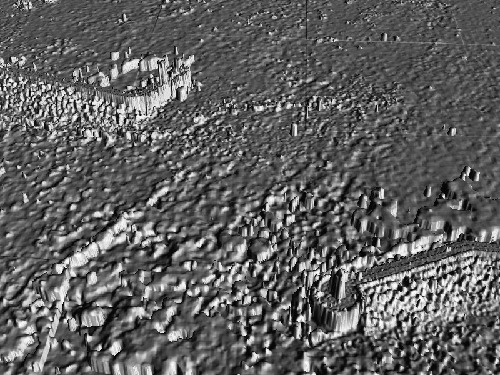}} \hfill
  \subfloat[]{\includegraphics[height=1.1in]{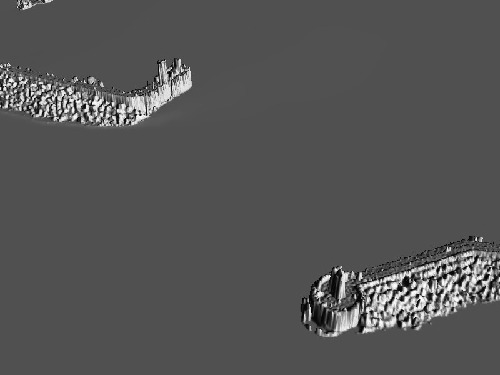}}
  \caption{(a) A 1~km wide inset of an imagery collection on the north coast of France. (b) 2D water classification mask for this area. (c) 3D reconstruction of the jetties without water classification. (d) 3D reconstruction of the jetties with water classification.}
  \label{fig:results_omahabeach}
\end{figure*}

\begin{figure*}
  \centering
  \subfloat[]{\includegraphics[height=1.1in]{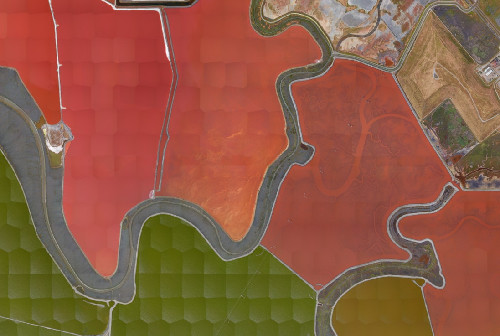}} \hfill
  \subfloat[]{\includegraphics[height=1.1in]{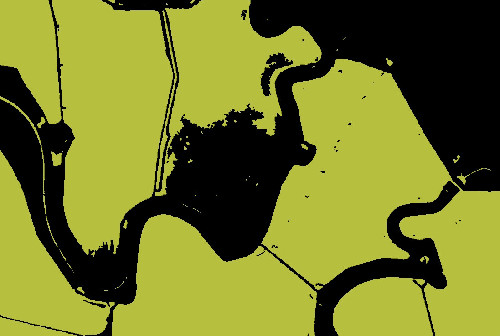}} \hfill
  \subfloat[]{\includegraphics[height=1.1in]{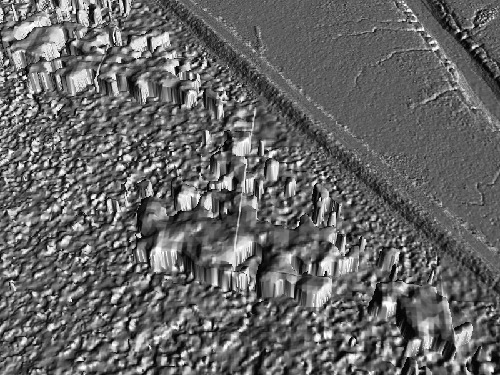}} \hfill
  \subfloat[]{\includegraphics[height=1.1in]{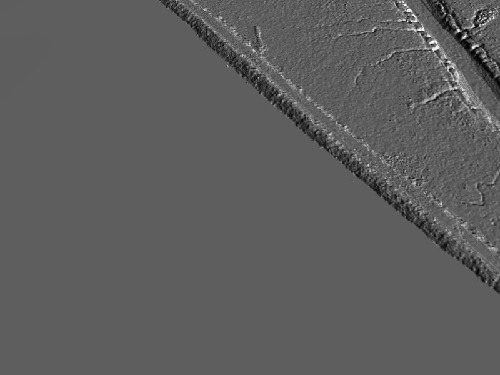}}
  \caption{(a) A challenging area that is quite different from our training data. This 4~km wide part of San Francisco Bay contains salt ponds with narrow roads between them. Micro-algae in the water make the coloration unusual. (b) 2D water classification mask for this area. Note that the mask ends where the water becomes mud. Also, we are able to recover fine details such as the pillars of electrical towers. (c) Detail showing a 200~m wide region of our 3D reconstruction, without applying the water mask. (d) Reconstruction after applying the water mask.}
  \label{fig:results_saltpond}
\end{figure*}

We used our classification system to implement an end-to-end system to find and fill water gaps in a world-scale 3D reconstruction task. See Figures~\ref{fig:results_niagara}--\ref{fig:results_saltpond} for examples of our water detection and filling algorithms running at scale on various data sets.

\subsection{Classifying trees}

To demonstrate the generality of our approach, we also applied our technique to classify trees. As with the water classification, we avoided onerous labelling of individual images by drawing polygons in all-tree and no-tree parts of the map, and using any imagery within those regions for training. Eight such polygons were used in this experiment.

\begin{figure}[hbt]
  \centering
  \subfloat[]{\includegraphics[scale=0.35]{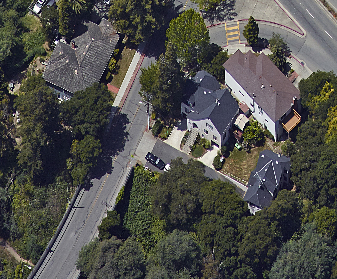}} \hfill
  \subfloat[]{\includegraphics[scale=0.35]{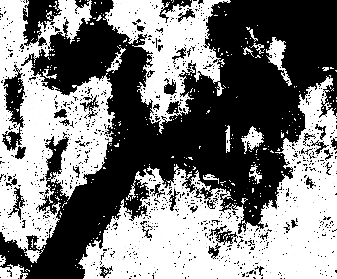}}
  \caption{Our approach used to classify trees. An example image is shown in (a) and the resulting classification in (b).}
  \label{fig:tree_image}
\end{figure}

Figure~\ref{fig:tree_image} demonstrates how our trained model works on a source image with both tree and non-tree regions.

\section{Conclusions}

In this paper, we presented a method for multi-view 3D classification. Motivated by reconstruction of aerial imagery, we extended this method to perform a consistent 2D classification. While the method is general, we explored it in the context of two applications of high practical value: classifying water, and classifying trees. For the example of water, we added a technique to fill gaps. We demonstrated the viability of our method with large-scale experimental results on real-world data. We also demonstrated by comparison that the multi-view nature of our setup is necessary to achieve these results.

\FloatBarrier

{\small
\bibliographystyle{ieee}
\clearpage
\newpage
\bibliography{vol_water_class}
}

\end{document}